\documentclass[10pt, logo, copyright]{nvidiatechreport}

\usepackage{mdframed}
\usepackage[utf8]{inputenc} 
\usepackage[T1]{fontenc}    

\usepackage{amsfonts}       
\usepackage{nicefrac}       
\usepackage{microtype}      
\usepackage[dvipsnames]{xcolor}         
\usepackage{multirow}
\usepackage{multicol}
\usepackage{graphicx}
\usepackage[numbers]{natbib}
\usepackage{tabto}
\usepackage{xspace}
\usepackage{amsmath}
\usepackage{adjustbox}
\usepackage{enumitem}
\usepackage{wrapfig}
\usepackage{dblfloatfix}

\usepackage{footmisc}
\usepackage{listings}

\usepackage{float}

\usepackage{hyperref}
\usepackage{array}
\usepackage{ragged2e}
\usepackage{subcaption}
\usepackage{caption}

\usepackage{natbib}

\usepackage{amsmath}
\usepackage{amssymb}
\usepackage{mathtools}
\usepackage{amsthm}
\usepackage{enumitem}
\setlist[itemize]{itemsep=0.5em}
\usepackage{multirow}
\usepackage{booktabs}

\usepackage[capitalize,noabbrev]{cleveref}

\usepackage{algorithm}
\usepackage{algorithmicx}
\usepackage{algpseudocode}
\usepackage{xspace}
\usepackage{csquotes}
\usepackage{listings}
\usepackage{xcolor}
\definecolor{NvidiaGreen}{rgb}{0,0.392,0}

\usepackage{multirow}
\usepackage{multicol}
\usepackage{enumitem}
\usepackage[normalem]{ulem}
\usepackage{pifont}
\usepackage{tikz}
\usetikzlibrary{calc}
\DeclareRobustCommand{\halfCmark}{%
  \tikz[baseline=(c.base)]{
    \node[inner sep=0pt] (c) {\ding{51}};
    \draw[line width=0.5pt] 
      ($(c.north)+(0.1ex,0)$) -- ($(c.east)+(0,-0.1ex)$);
  }%
}

\DeclareRobustCommand{\ours}{{TiDAR}}

\usepackage[most]{tcolorbox}

\definecolor{lightgray}{gray}{0.95} 
\definecolor{darkblue}{rgb}{0,0,0.6} 
\definecolor{nvgreen}{cmyk}{50, 0, 100, 0}

\title{TiDAR: Think in Diffusion, Talk in Autoregression}

\author{Jingyu Liu\textsuperscript{*}\textonesuperior, Xin Dong\textsuperscript{*}, Zhifan Ye\texttwosuperior, Rishabh Mehta, Yonggan Fu, Vartika Singh, Jan Kautz, \space \space Ce Zhang\textonesuperior, Pavlo Molchanov  \\~\\ NVIDIA \\ \textsuperscript{*}Equal Contribution; XD as project lead; Corresponding: jingyu6@uchicago.edu, xind@nvidia.com \\~\\~\\ }
\correspondingauthor{Xin Dong}
\begin{abstract}
\textbf{Abstract:} Diffusion language models hold the promise of fast parallel generation, while autoregressive (AR) models typically excel in quality due to their causal structure aligning naturally with language modeling. This raises a fundamental question: can we achieve a synergy with high throughput, higher GPU utilization, and AR level quality?
Existing methods fail to effectively balance these two aspects, either prioritizing AR using a weaker model for sequential drafting (speculative decoding), leading to lower drafting efficiency, or using some form of left-to-right (AR-like) decoding logic for diffusion, which still suffers from quality degradation and forfeits its potential parallelizability.

We introduce \ours{}, a sequence-level hybrid architecture that drafts tokens (\textbf{T}hinking) \textbf{i}n \textbf{D}iffusion and samples final outputs (Talking) \textbf{A}uto\textbf{R}egressively - \textit{all within a single forward pass using specially designed structured attention masks}. This design exploits the free GPU compute density, achieving a strong balance between drafting and verification capacity. Moreover, \ours{} is designed to be serving-friendly (low overhead) as a standalone model. 

We extensively evaluate \ours{} against AR models, speculative decoding, and diffusion variants across  generative and likelihood tasks at 1.5B and 8B scales. 
Thanks to the parallel drafting and sampling as well as exact KV cache support, \ours{} outperforms speculative decoding in measured throughput and surpasses diffusion models like Dream and Llada in both efficiency and quality. Most notably, \ours{} is the first architecture to close the quality gap with AR models while delivering 4.71× to 5.91× more tokens per second. 
\end{abstract}

\begin{document}
\maketitle

\section{Introduction}
\label{introduction}
As we move towards Artificial General Intelligence~\cite{bubeck2023sparks}, the remarkable success of large language models (LLMs) can be largely attributed to their ability to harness the massive computational scaling afforded by the explosively increasing power of GPUs~\cite{nvidia_gpu}.
So how to make the most of compute resources during both training and testing times becomes increasingly important. Although autoregressive models~\cite{vaswani2017attention,radford2019language} are the prevailing approach, they remain memory-bound during decoding~\cite{dao2022flashattention, yuan2024llm} and cannot fully exploit hardware compute density, particularly at small batch sizes, since they generate only one token per step. In contrast, diffusion language models (dLMs)~\cite{sahoo2024simpleeffectivemaskeddiffusion,nie2025large,ye2025dream} offer the promise of parallel token decoding. However, they typically face a trade-off between quality and parallelizability.
In this work, we provide a principled analysis to identify where these models fall short and propose a simple yet effective hybrid architecture that combines the strengths of both paradigms.

\begin{figure}[h]
    \centering
    \includegraphics[width=0.5\columnwidth]{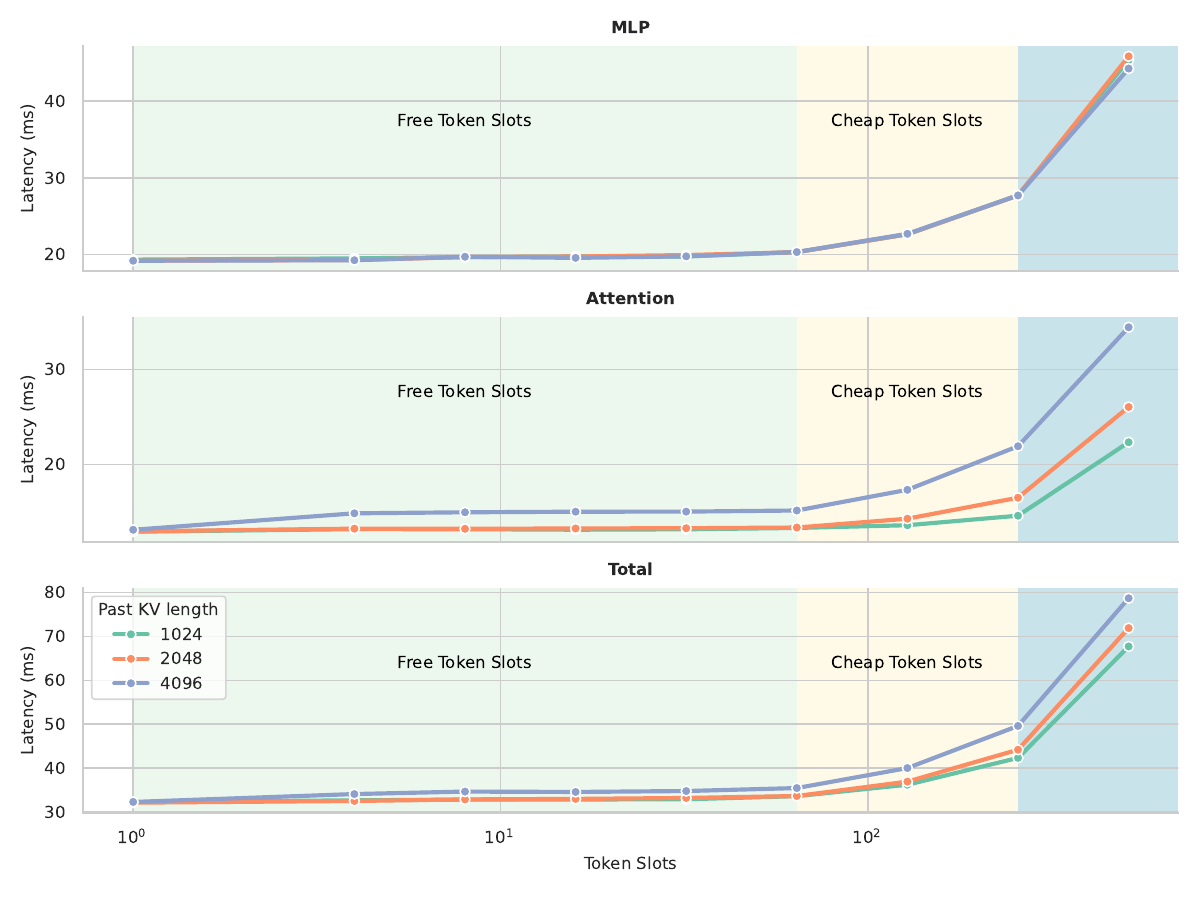}
    \vspace{-0.2cm}
    \caption{\textbf{Latency Scaling over Token Slots:} We plot the latency of Qwen3-32B decoding on NVIDIA H100 with batch size=1 and Flash Attention 2~\cite{dao2023flashattention2fasterattentionbetter} over different prefix lengths. Latency stays relatively the same with a certain amount of tokens sent to forward (free + cheap token slots), before transitioning to the compute-bound regime. We leverage this characteristic to achieve almost free parallelled drafting and sampling for \ours{}. }
    \label{fig:latency_scaling}
\end{figure}

Decoding in AR models is memory-bound because the latency is dominated by loading the model weights and KV cache rather than compute~\cite{kwon2023efficient, leviathan2023fast}. If we can decode multiple tokens in a single forward pass, those tokens can share the same loaded weights and KV cache, increasing the compute density without increasing end-to-end latency before transitioning to the compute-bound region. This is precisely why dLMs could provide potential speedup from a systems perspective. Concretely, given a prefix $x_{<t}$ and model function $F$, an AR model appends a single token slot (last prompt token) and predicts one next token, e.g., $x_{t + 1} := F([x_{<t-1};x_{t}])$. In a masked diffusion model (e.g., Block Diffusion~\cite{arriola2025blockdiffusion} with a simplified one-step denoising scenario), instead we predict multiple tokens at once, $x_{t+1}, \dots, x_{t+k+1} := F([x_{<t};m_{t + 1}, \dots, m_{t+k+1}])$ (due to no label shifts). If for a given $k$ such that both computations are still memory-bound, the forward time of $F([x_{<t-1};x_{t}])$ and $F([x_{<t};m_{t+1}, \dots, m_{t+k+1}])$ should be similar. We refer to the extra token slots $(m_{t+2}, \dots, m_{t+k+1})$ as \textbf{free token slots} because carrying them through a single forward incurs minimal to no latency increase as validated by a real-world profiling in Figure~\ref{fig:latency_scaling}.

However, a well-known tension between parallel decoding and output quality exists for masked diffusion models~\cite{wu2025fastdllmtrainingfreeaccelerationdiffusion, feng2025theoretical}: the best quality is often achieved when decoding strictly one token per denoising step, while attempting to exploit within-step parallelism tends to degrade quality. Consequently, current open source SOTA diffusion LLMs, such as Dream~\cite{ye2025dream} and Llada~\cite{nie2025large}, have not yet matched the combined speed–quality profile of strong AR LLMs. To formalize their difference from a modeling perspective, AR models sample from a chain-factorized joint distribution:
\begin{equation*}
    p_{\text{AR}}(\cdot; \theta) = \prod_i p^i_\theta(x_i | \mathbf{x}_{<i}; \theta).
    \label{equ:p_joint}
\end{equation*}
A diffusion model, on the other hand, samples from the following distribution: 
\begin{equation*}
p_{\text{Diff}}(\cdot;\theta) = \mathbb{E}_{\tilde{\mathbf{x}} \sim q(\cdot | \mathbf{x})}\prod_i p^i_\theta(x_i | \tilde{\mathbf{x}})
\end{equation*}
Often the quality is best preserved if we choose to decode one token $x_i$ instead of $k$ tokens $\{x_j, j \in K\}$ per step because in this case $\tilde{\mathbf{x}}$ will have the decoded token as a condition for the next denoising step. Decoding $k$ tokens in one step will further factorize $p_\text{Diff}$ as a product of marginals:

\begin{equation*}
p_{\text{Diff\_Independent\_K}}(\cdot;\theta) = \mathbb{E}_{\substack{\tilde{\mathbf{x}} \sim q(\cdot | \mathbf{x}) \\ s.t.\space \tilde{\mathbf{x}}_{i\in K}=[m]}}\prod_{i\in K} p^i_\theta(x^i_t | \tilde{\mathbf{x}})
\end{equation*}

The introduced token independence assumption will degrade the quality despite providing more parallelism during sampling. The diffusion sampling procedure would fall back to AR if we restrict the order as strict left-to-right, decode one token per step, and change the masking strategy $q(\cdot | \mathbf{x})$ to uniform suffix masking during training.  


Ideally, we hope to compute with $p_{\text{Diff}}$ and leverage its independence assumption for parallel sampling, but, at the same time, we want to get the quality from $p_{\text{AR}}$ due to its aligned casual factorization nature to language modeling and more context in the condition. 

To this end, we propose \ours{}, a new architecture that enables parallel token computation (``thinking'') from the marginal distribution via diffusion, and high-quality sampling (``talking'') from the chain-factorized joint distribution via autoregression. 
Concretely, at each generation step (one forward pass), we partition tokens into three sections: prefix tokens, tokens proposed in the previous step, and tokens pre-drafted for the next step. 
We reuse the KV cache of prefix tokens from the last step. 
Tokens proposed from the last step are autoregressively sampled via rejection sampling guided by $p_\text{AR}$ computed at current step. 
At the same time, we pre-draft proposals from the $p_\text{Diff}$ conditioned on all possible prefix outcomes of the rejection sampling.  
One pre-draft proposal will be chosen from them, and passed to the next step. 
All of these happen in a single forward pass with a simple and well-designed attention mask.
The overhead is almost negligible as long as the tokens residing in the last two sections can be well fitted into the ``free token slots''.
Figure~\ref{fig:main} illustrates the architecture in detail.


Training \ours{} is straightforward and data-efficient, as it employs a structured causal–bidirectional hybrid attention mask over the input sequence, enabling it to learn and sample from $p_\text{AR}$ and $p_\text{Diff}$ within a single model and foward.
Consequently, both autoregressive and diffusion losses can be computed on the same data sample.
During training, all tokens in the diffusion section are set to mask tokens. This simplifies the masking strategy, strengthens the diffusion loss signal, promotes train-test consistency, and balances the autoregressive and diffusion objectives.

\textbf{Our contributions can be summarized as: }

\begin{itemize}
    \item We propose a sequence-level hybrid architecture called \ours{} that utilizes the ``free token slots'' to conduct parallel token drafting via diffusion and sampling via autoregression, which combines the speed and quality advantages from both paradigms.  
    \item We provide a training recipe as well as comprehensive evaluations on both likelihood and generative downstream tasks to support its superiority over other architectures. 
    \item We conducted detailed ablations to prove the effectiveness of our core design choices and flexibility. In addition, we analyze \ours{} from the lens of diffusion and speculative decoding, providing a clear understanding of why our method is appealing. 
    \item We show that for \ours{} 1.5B, we can achieve lossless quality compared to its AR counterpart while generating with 4.71$\times$ relative throughput (tokens per second) speedup. For \ours{} 8B, we achieved an impressive 5.91$\times$ relative throughput speedup with minimal loss. 
\end{itemize}


\section{Background and Related Work}
\label{related_works}

\begin{table}[h]
\centering
\fontsize{7}{9}\selectfont
\begin{tabular}{lcc|cc}
\toprule
 & \multicolumn{2}{c}{\textbf{Draft Model}} & \multicolumn{2}{c}{\textbf{Drafting Process}} \\
\midrule
\textbf{Model} & \textbf{Shared with Base} & \textbf{Drafting Capacity} & \textbf{Parallel Decoding} & \textbf{Parallel to Verification} \\
Classic Spec. Decoding~\cite{chen2023accelerating} & \ding{55} & Low & \ding{55} & \ding{55} \\ 

\multirow{2}{*}{APD~\cite{israel2025apd}} & \multirow{2}{*}{\ding{55}} & High & \multirow{2}{*}{\ding{51}} & \multirow{2}{*}{\ding{55}} \\
 & & (Weak Verifier) & & \\

EAGLE-3 \cite{li2025eagle3} \& & \multirow{2}{*}{\halfCmark} & \multirow{2}{*}{Mid} & \multirow{2}{*}{\ding{55}} & \multirow{2}{*}{\ding{55}} \\
DeepSeek-V3 \cite{liu2024deepseekv3} & & & & \\ \addlinespace
Apple MTP~\cite{samragh2025llmknowsfutureuncovering} & \halfCmark & Mid & \ding{55} & \ding{51} \\
\midrule
\ours{} & \ding{51} & High & \ding{51} & \ding{51} \\
\bottomrule
\end{tabular}
\caption{\textbf{Comparison among Speculative Frameworks:} We compare different speculative frameworks in two major aspects: 1) whether the drafter is a separate module from the base model and has high capacity. 2) if the drafting process has paralleled decoding and is sequential to the verification or sampling process. \halfCmark\ means that draft models need extra layers and heads to the base model.}
\label{tab:compare_sd}
\end{table}

\ours{} model is related to and enjoys the benefits of two lines of works: diffusion language models and speculative decoding. In this section, we connect \ours{} to these two categories and highlight how we understand \ours{} from different perspectives and how it improves upon prior works. 

\subsection{Diffusion Language Models}

Diffusion language models~\cite{austin2021structured,li2022diffusion,nie2025large, ye2025dream, arriola2025blockdiffusion,sahoo2024simpleeffectivemaskeddiffusion} (dLMs) offer a promising alternative to purely sequential generation of autoregressive models by allowing parallel generation of multiple tokens at each step, offering a path towards significant speed-up in generation. Although theoretically appealing, dLLMs face a trade-off contradiction between parallelizability and generation quality. As exemplified by open-source dLLMs like Llada~\cite{nie2025large} and Dream~\cite{ye2025dream}, the best generation quality is often achieved when decoding one token at a time (1 token per model function forward).
As reported in APD~\cite{israel2025apd}, there is a clear trend of declining generation quality with increasing number of tokens to generate in parallel per step. Specifically, the accuracy on GSM8K~\cite{cobbe2021training} drops by 10\% when increasing from 1 to 2 tokens per step for Dream-7B with the entropy-base sampling strategy~\cite{ye2025dream}. When decoding multiple tokens per step, dLMs typically sample each token independently from the marginal distribution, which introduces an intra-step token independence assumption~\cite{wu2025fastdllmv2efficientblockdiffusion} that can hurt sequence-level coherence and correctness~\cite{feng2025theoretical}. 
Recent works have attempted to address the gap between the generation quality of diffusion models and AR and improve the throughput-quality trade-off. E2D2 ~\cite{arriola2025encoderdecoderdiffusionlanguagemodels} adopts an encoder-decoder architecture that enables disaggregating processing FLOPs by token type, where a large encoder processes clean tokens and a lightweight decoder is tasked with decoding noisy tokens. EDLM ~\cite{xu2025energybaseddiffusionlanguagemodels} introduces residual energy-based approach to reduce mismatch between training and sampling distributions, improving generation quality.

Another challenge of scaling up dLM is the lack of support for exact KV caching, as a result of bidirectional attention. 
Fast-dLLM~\cite{wu2025fastdllmtrainingfreeaccelerationdiffusion,wu2025fastdllmv2efficientblockdiffusion} proposes to perform block parallel decoding with the prefix and optionally the suffix being cached during the process of denoising the current block. d-KV cache~\cite{ma2025dkvcachecachediffusionlanguage}, on the other hand, takes a more dynamic route by selectively cache certain tokens step by step in a delayed fashion to minimize quality degradation. 
Block Diffusion (also known as semi-autoregressive models)~\cite{arriola2025blockdiffusion} attempts to address this issue by interpolating between discrete diffusion and autoregressive models. Specifically, the model defines an autoregressive probability distribution across blocks and the conditional probability of tokens within a block given previous blocks is specified by a denoising discrete diffusion model. Despite being equipped with exact caching, Block Diffusion still suffers from the same dilemma of quality degradation and intra-block token parallelizability. 

\subsection{Speculative Decoding}
\ours{} is also closely related to speculative decoding~\cite{chen2023accelerating}, which accelerates generation by first using a faster draft model to generate a sequence of candidate tokens and then uses a modified rejection sampling strategy~\cite{leviathan2023fast} to validate the these tokens against the target distribution of the base model. In order to improve the speed of drafting, a smaller draft model is usually used. However, if degradation in drafting quality is too severe, the overall generation speed can be slowed down because of low acceptance rate of draft. In order to increase both the speed and the quality of drafting, prior works propose to carry the hidden states from the base model to the drafter. The intuition is that the model's hidden states (e.g., the second-to-top layer embedding) carry richer information not only for the next token prediction but also for future tokens. For example, Medusa~\cite{cai2024medusa} adds extra multiple linear decoding heads on the base model's hidden states to predict future tokens with an efficient tree verification pattern to expand possible paths. EAGLE series~\cite{li2024eagle1,li2024eagle2,li2025eagle3} and DeepSeek-V3 Multi-Token Prediction (MTP)~\cite{liu2024deepseekv3} use additional autoregressive layers on the base model's hidden states to predict future tokens. 
However, the drafting process does not fully take advantage of the base model, and the maximal speedup is hindered by the lower drafter capacity. 
In addition, EAGLE and DeepSeek-V3's MTP modules are still autoregressive and sequential to the base verification. These two factors show that they cannot effectively increase the compute density and release the full power of parallel generation. 

If we view \ours{} from the perspective of speculative decoding (summarized comparison in Table~\ref{tab:compare_sd}), one of the main advantages of \ours{} is that it only has a single model and can complete both the drafting and sampling simultaneously in a single forward pass. This induces three merits: 
1) The draft model is the base model itself, so it has high capacity by reusing the base model's weights and compute. 
2) The drafting process follows a diffusion approach, allowing it to be fully parallelized. This means it can utilize the full computational power across all input mask tokens, rather than being limited to only the last token as in autoregressive MTP methods.
3) The drafting and verification processes are parallelized in a single forward pass, which further eliminates the overhead of sequential drafting and verification.

\section{Method}
\label{method}

\begin{figure}[t]
    \centering
    \includegraphics[width=0.8\textwidth]{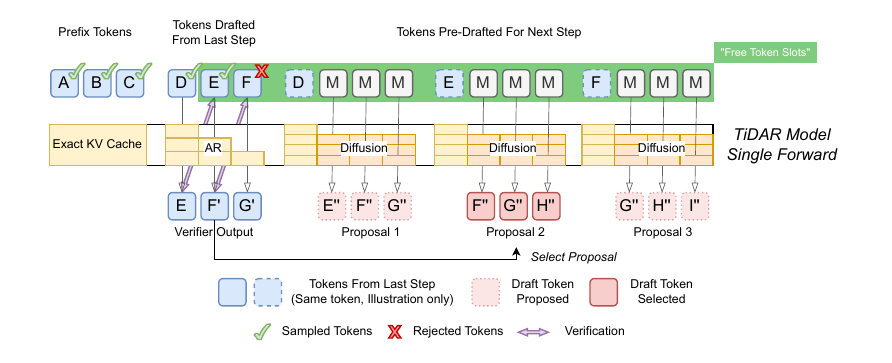}
    \caption{\textbf{\ours{} Architecture:} \ours{} uses a single model forward to sample drafted tokens from the last step and pre-draft tokens for the next step in parallel. By switching the attention pattern among different parts of the sequence, \ours{} encodes the clean tokens drafted from last step causally and mask tokens block-causally (bidirectional within each block) for one-step diffusion pre-drafting. Upon accepting a prefix, the corresponding pre-drafts (proposal) can be selected. The KV cache for tokens forwarded causally will be stored and later evicted if the corresponding tokens are rejected. We illustrate this with a draft length of 3 and an accepted length of 2. Figure~\ref{fig:mask} shows the exact decoding mask for this example. }
    \label{fig:main}
\end{figure}

\begin{figure}[!h]
    \centering
    \includegraphics[width=0.8\textwidth]{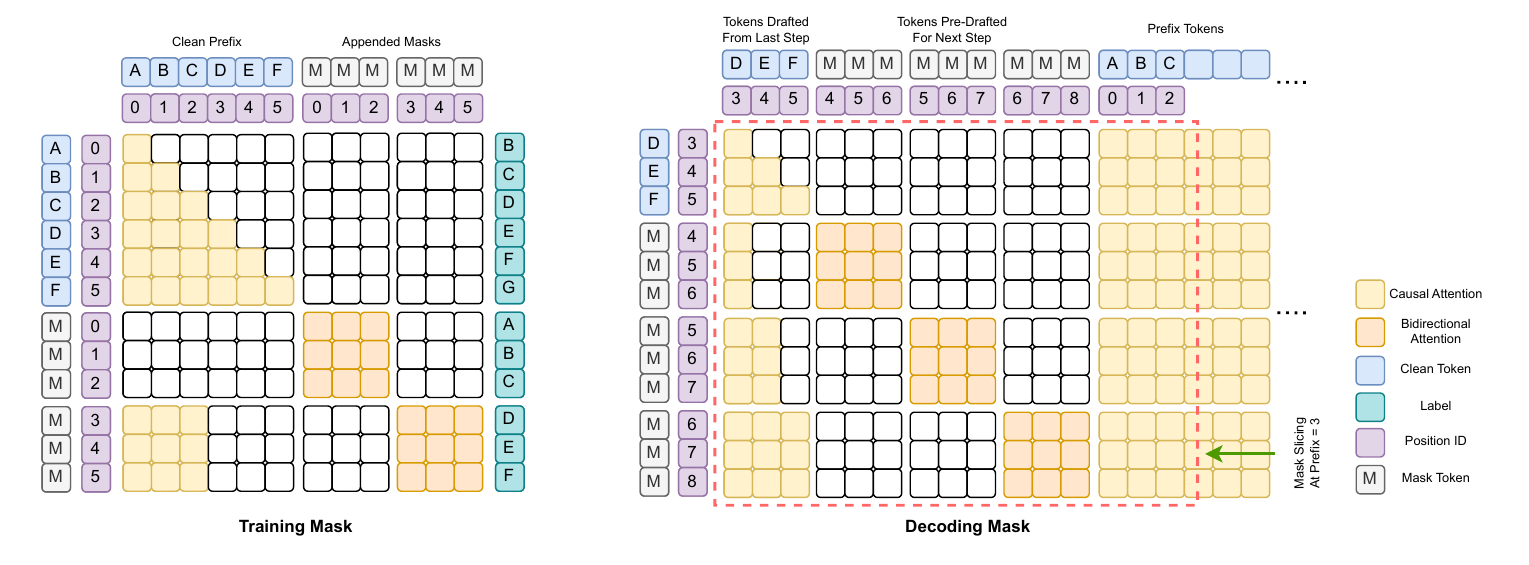}
    \caption{\textbf{\ours{} Attention Masks:} (Left) We apply a special training mask (using block length = 3): mask tokens of the same length are appended to the input tokens where the clean input tokens are self-attended causally and mask tokens within-block bidirectionally along with the prefix. During inference parallel decoding, we use a slice of a pre-initialized mask based on the prefix of the current step (Right). To reuse the mask, we reorder the sampling-draft part (tokens drafted from last step and mask positions for next step pre-drafting) and the clean prefix as illustrated with an example prefix length of 3. }
    \label{fig:mask}
\end{figure}

In this section, we describe the proposed architecture \ours{}, starting from training the backbone that enables modeling both 
the joint distribution (i.e., AR mode) and the marginal distribution (i.e., diffusion mode) in a single model and a single model forward. We then elaborate how we use the dual modes for paralleled self-speculative generation, allowing the model to draft (``thinking'') in diffusion for efficiency and sample (``talking'') in autoregression for quality. Finally, we conclude by discussing the training and inference optimizations.

\subsection{Diffusion-AR Dual-mode Backbone Training}
\label{sec:backbone}
Our goal is to train a model that enables the diffusion and AR dual modes. More importantly, we want to have the dual modes happening in a single model forward in order to exploit the ``free token slots''. To achieve this, sequence level casual and bidirectional attention hybridization becomes most natural choice. 
A similar idea has been explored in the context of Block Diffusion models~\cite{arriola2025blockdiffusion}; however, their primary objective is not to support dual-mode operation, but rather to enable KV caching and improve generation quality specifically for the diffusion component. The resulting attention mask of Block Diffusion is intra-block bidirectional and inter-block causal.
We make a modification to Block Diffusion by only preserving the last block, which is the decoding block, to be bidirectional and the rest (i.e. prefix) to be causal. 
The benefits of doing so are two-folds. 
First, it allows us to compute the chain-factorized joint distribution just like in AR models. This, as we will show in Section~\ref{sec:generation}, will allow us to conduct rejection sampling using the joint distribution with high quality guarantee and evaluate likelihood the same way as AR in terms of efficiency (Section~\ref{sec:likelihood}). 
Second, computing next token prediction (NTP) loss on the prefix becomes possible during model pre-training and finetuning~\cite{gat2025set}. 
Note that Block Diffusion cannot compute this loss on the prefix because of the label leakage issue of intra-block bidirectional attention. 
In addition, the NTP signal is much denser than pure diffusion loss since the latter is only applied to the mask tokens. In our model training, we can leverage the two types of losses simultaneously to better utilize the training data. 

In Figure~\ref{fig:mask} (Left), we visualize the attention masked during training. 
Similar to Block Diffusion~\cite{arriola2025blockdiffusion} and Set Block Decoding (SBD)~\cite{gat2025set}, we also need to double the sequence length, as a result of appending original input sequence with corrupted tokens.  
For the causal (autoregressive) section, the target labels are shifted by one position to match NTP objective, while for the bidirectional (diffusion) section, the labels remain aligned with their input positions.
For the diffusion part, Block Diffusion~\cite{arriola2025blockdiffusion} and SBD~\cite{gat2025set} add noise to the input sequence by applying random masks sampled from a special distribution, following the traditional diffusion language modeling approaches~\cite{nie2025large,ye2025dream,sahoo2024simple}. 

We propose a simpler and more effective training strategy by setting all tokens in the diffusion section to mask tokens. This eliminates the hassle of deciding the optimal masking strategy. And more importantly, it allows us to compute the loss for every token in the diffusion part, resulting in three key benefits: (1) The diffusion loss becomes much denser compared to only corrupted tokens like before. (2) Balancing the diffusion loss with the NTP loss becomes much simpler. In traditional diffusion approaches, the number of loss terms varies across samples depending on how many tokens are masked, whereas the next-token prediction loss is always calculated over the same number of tokens. This mismatch makes it difficult to balance the two losses. By masking all tokens, the number of loss terms is consistent across both types of losses (equal to the sequence length), allowing straightforward balancing using a user-defined weighting factor. (3) It allows us to do one-step diffusion during inference, which makes the drafting process more efficient than multi-step denoising. We detail this with ablation studies in Section~\ref{sec:full_mask}.

Therefore, the training objective of \ours{} is simplified as: 

\begin{align*}
    \mathcal{L}_{\ours{}}(\theta) = \frac{1}{1 + \alpha} \bigg( \sum^{S - 1}_{i=1} \frac{\alpha}{S - 1}  \cdot \mathcal{L}_{AR}(x_i, x_{i + 1}; \theta) \\
    +  \sum^{S - 1}_{i=1} \frac{1}{S - 1} \cdot \mathcal{L}_{Diff}([mask], x_i; \theta) \bigg)
\end{align*}

where $\alpha\in[0, 1]$ is the loss balancing factor, $\{x_i\}_S$ is the input sequence with length $S$, and $\mathcal{L}_{AR}, \mathcal{L}_{Diff}$ are cross-entropy losses with logits calculated at clean and masked sequences using different attention patterns. 

\subsection{Fully Parallelizable Self-Speculative Generation}
\label{sec:generation}

Reasoning the generation efficiency of a model requires considering many aspects: the number of forward steps in total, the number of decoded (or sampled) tokens per step, as well as the latency per step. 
Many existing techniques falling in the category of diffusion language models~\cite{nie2025large, ye2025dream,israel2025apd}, MTP~\cite{gloeckle2024better,cai2024medusa,liu2024deepseekv3}, and speculative decoding~\cite{leviathan2023fast,li2024eagle1} are trying to optimize some of them. However, the result is often a trade-off among latency, quality, and compute density.  
In this work, we attempt to provide an answer to this global optimization problem by incorporating the diffusion parallelism and AR quality into a single model. 

We propose a parallel drafting and sampling procedure. The method centers around the speculative framework where the model first drafts speculative tokens in parallel from the marginal distribution, which are then rejectively sampled in an autoregressive manner to secure generative quality. 
In the start of the generation, the model encodes prompt causually and drafts a block of tokens in parallel with a bidirectional attention (mask illustrated in Appendix Figure~\ref{fig:prefill}). 
In the each subsequent decoding step, draft tokens from the last step are rejectively sampled by checking whether they match the prediction from the autoregressive joint distribution computed at current step using causal attention. 
In the same time, inspired by Apple's MTP work~\cite{samragh2025llmknowsfutureuncovering}, we also pre-draft the next step's tokens in parallel from the marginal distribution using bidirectional attention, conditioned on all possible outcomes of the rejection sampling.
So that no matter how many tokens we accept at the current step, we would be able to get the corresponding drafts for the next step. 
In Figure~\ref{fig:main}, we illustrate the generation process. 


To further improve the efficiency, we adopt a one-step diffusion drafting~\cite{liu2025sequential}. We found one step is sufficient to produce draft tokens whose quality is good enough to secure high acceptance rate. Correspondingly, we set all tokens in the diffusion section to masks during training as discussed in the above Section~\ref{sec:backbone}.

\subsection{Training and Inference Optimization}
\label{sec:optimization}
As we discuss in Section~\ref{sec:backbone}, our corrupted sequence is fully masked, which makes loss balancing easier due to the equal variance of the loss scale calculated from the AR and diffusion logits. We choose to set $\alpha=1$ for most cases and ablate the difference of different $\alpha$ values in Section~\ref{sec:sampling_ar_vs_diff}. 

During inference, we leverage the fact that the number of tokens we send to the model in each forward is the same and has the same attention pattern, we reorder the draft part and prefix so that we can initialize one block attention mask of size (q\_len\footnote{q\_len = block\_len $\cdot$ (1 + block\_len)}, q\_len + max\_sequence\_len) and slice the cached mask for each step without recomputing it for Flex Attention~\cite{dong2024flex} (Figure~\ref{fig:mask} (Right)). Different from traditional diffusion models, \ours{} supports exact KV cache like Block Diffusion. In Figure~\ref{fig:main}, we showcase the generative process where we first save all the KV cache of tokens which are computed with causal attention, and later on, evict the KV cache if our sampling length is shorter than the draft length. It is worth mentioning that we do not waste any computation by recomputing the KV cache of any token, which makes our method extremely efficient compared to Block Diffusion, SBD, and the cache methods used in pure diffusion (e.g. Fast-dLLMs~\cite{wu2025fastdllmtrainingfreeaccelerationdiffusion, wu2025fastdllmv2efficientblockdiffusion} and d-KV Cache~\cite{ma2025dkvcachecachediffusionlanguage}). 

Unlike traditional diffusion models, \ours{} has no hyperparameters to tune during inference. But still, we provide some flexibility to accomondate different scenarios as detailed in Section~\ref{sec:compare_decoding} and ~\ref{sec:sampling_ar_vs_diff}. 

\begin{figure}[t]
    \centering
    \includegraphics[width=\textwidth]{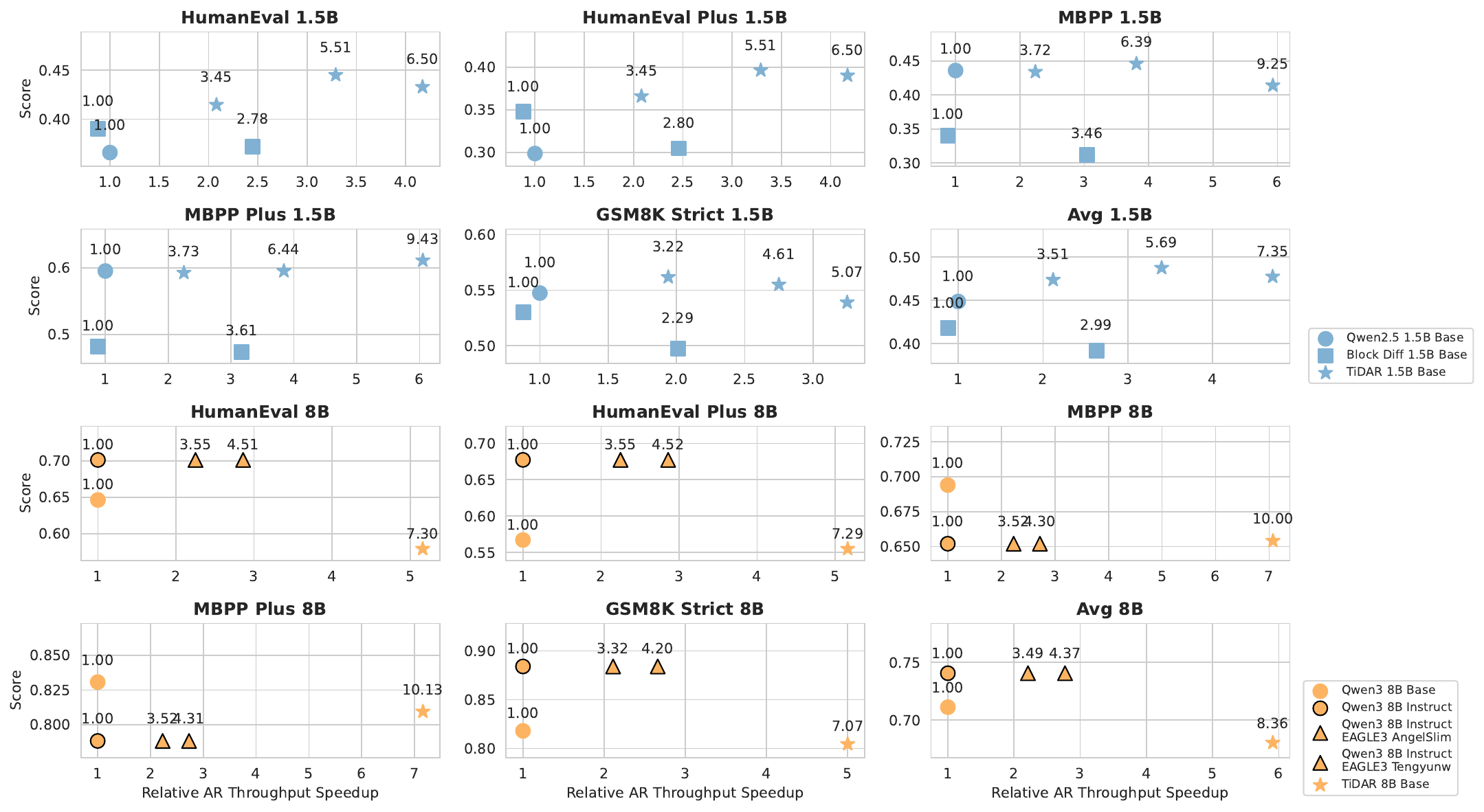}
    \caption{\textbf{Efficiency-Quality Benchmarking:} We compare \ours{} on 1.5B and 8B with AR, AR with speculative decoding (EAGLE-3), and Block Diffusion. Points colored the same indicate the same model sizes while markers suggest different methods. On the y-axis we have individual task scores. On the x-axis, we showcase the relative decoding throughput speedup measured in tokens per second, with the baseline being the AR model within the same size group (\protect\tikz\fill[cyan] (0,0) circle (0.1cm); Qwen2.5 1.5B Base, \protect\tikz\fill[orange] (0,0) circle (0.1cm); Qwen3 8B Base and \protect\tikz\fill[orange, draw=black, thick] (0,0) circle (0.1cm); Qwen3 8B Instruct). On top of each point, we report the average tokens per NFE. For 1.5B models, we showcase two and three different settings for Block Diffusion (threshold = max, 0.8, illustrated from left to right) and \ours{} (training block size = 4, 8, 16, illustrated from left to right) respectively. }
    \label{fig:throughput}
\end{figure}

\section{Experiments}
\label{experiments}

\subsection{Setup}

\paragraph{Model initialization and tasks} In this paper, we focus on the setting of continual pretraining from AR models (Qwen2.5 1.5B~\cite{qwen2025qwen25technicalreport}, Qwen3 4B, and Qwen3 8B~\cite{yang2025qwen3technicalreport}). We include quality evaluations on generative and likelihood tasks, including coding (HumanEval, HumanEval+, MBPP, and MBPP+), math (GSM8K and Minerva Math), factual knowledge (MMLU), and commonsense reasoning (ARC, Hellaswag, PIQA, and Winogrande). We use \textsc{lm\_eval\_harness}~\cite{eval-harness} for all evaluation. The detailed task configurations can be found in the Appendix~\ref{sec:app_eval_task_config}. In terms of efficiency, we report the average number of tokens a single model forward (or network function evaluation, NFE for short) can produce, as well as wall-block speedup in terms of token per wall-clock second under the batch size of one.  


\paragraph{Baseline} We include open-sourced AR models such as Llama3.2~\cite{meta2024llama}, SmolLM2~\cite{allal2025smollm2}, Qwen2.5~\cite{qwen2025qwen25technicalreport}, and Qwen3~\cite{yang2025qwen3technicalreport} with similar model sizes. For diffusion models, we compare ours against Dream~\cite{ye2025dream} and Llada~\cite{nie2025large}, as well as our trained Block Diffusion~\cite{arriola2025blockdiffusion} under the same training recipe. \textit{For all Qwen2.5 and Qwen3 models, the results are referring to the base model unless explicitly specified. }

\paragraph{Training and inference} For \ours{} 1.5B model, we continually pretrain with 50B tokens on Nvidia H100s with a global batch size of 2M tokens (DDP) under block sizes of 4, 8, and 16. For the 8B model, we use 150B tokens instead, with a block size of 16, keeping the rest of the settings the same. We adopt the cosine scheduling with $max\_lr = 1e-5, min\_lr = 3e-6$ and a warm-up step fraction of $1\%$. The max sequence length is set to 4096 with distributed Adam~\cite{kingma2017adammethodstochasticoptimization} optimizer and we turn on gradient checkpointing for the 8B model. The overall training framework is a modified Megatron-LM~\cite{shoeybi2020megatronlmtrainingmultibillionparameter} with Torchtitan~\cite{liang2025torchtitanonestoppytorchnative} support. Both training and inference are conducted in the standard BFloat 16 precision. In terms of the main results, we report the performance for \ours{} 1.5B and \ours{} 8B and focus on the use of 1.5B for ablation studies.


\subsection{Main Results}

\subsubsection{Generative Task Evaluation}
In this section, we start off by investigating the ability of \ours{} to generate high-quality samples efficiently. We focus on coding and math tasks since their metrics are much more robust. In Table~\ref{tab:generative}, we first compare \ours{} against several AR models and also a popular diffusion variant, Block Diffusion, across two model sizes. 

For 1.5B-1.7B size range, \ours{} is highly competitive in terms of quality with an average 7.45 tokens per model forward (NFE). For 8B models, \ours{} incurs very minimal loss with a simple training recipe while increasing the generation efficiency to 8.25 tokens per NFE. We also tested two different modes of generation, namely ``trusting AR'' v.s. ``trusting diffusion'', which shows that for larger models, trusting the predictions from diffusion predictions is beneficial in most cases, especially for math tasks. We will discuss the choice difference more in Section~\ref{sec:sampling_ar_vs_diff}. 

Comparing with diffusion LLMs, \ours{} outperforms both public models like Dream and Llada consistently, and also our own Block Diffusion with the same training recipe. Here, we decode one token per NFE since this very often guarantees the best achievable quality for most tasks. We defer the discussion and comparison of different decoding strategies to Section~\ref{sec:compare_decoding}. 

In summary, \ours{} strikes a nice balance between generative quality and efficiency due to several designs we mentioned above, making it a highly appealing choice for many critical application scenarios with a stringent latency requirement. 

\begin{table}[h]
\centering
\fontsize{7}{9}\selectfont
\begin{tabular}{llcccc|cc|c}
\toprule
 & & \multicolumn{4}{c}{\textbf{Coding}} & \multicolumn{2}{c}{\textbf{Math}} & \multicolumn{1}{c}{\textbf{Avg}} \\
\midrule
\textbf{Model Arch} & \textbf{Size} & \multicolumn{1}{c}{\textbf{HumanEval}} & \multicolumn{1}{c}{\textbf{HumanEval+}} & \multicolumn{1}{c}{\textbf{MBPP}} & \multicolumn{1}{c}{\textbf{MBPP+}} & \multicolumn{1}{c}{\textbf{GSM8k}} & \multicolumn{1}{c}{\textbf{Minerva Math}} & \multicolumn{1}{c}{} \\
\midrule
Llama3.2 & 1B & 17.68\% & 14.63\% & 26.60\% & 38.89\% & 5.67\% & 1.92\% & 17.57\% \\
SmolLM2 & 1.7B & 0.61\% & 0.61\% & 35.40\% & 47.62\% & 28.20\% & 11.28\% & 20.62\% \\
Qwen2.5 & 0.5B & 27.44\% & 25.61\% & 29.60\% & 44.97\% & 37.45\% & 14.48\% & 29.92\% \\
Qwen2.5 & 1.5B & 35.98\% & 29.88\% & 43.60\% & 59.23\% & 54.74\% & 26.40\% & 41.64\% \\
Qwen3 & 1.7B & 48.17\% & 41.46\% & 55.80\% & 71.43\% & 66.72\% & 29.74\% & 52.22\% \\
\midrule
Block Diff & 1.5B$^{\star}$ & 39.02\% & 34.76\% & 34.00\% & 48.15\% & 52.99\% & 21.56\% & 38.41\% \\
\multirow{2}{*}{\ours{}} & \multirow{2}{*}{1.5B$^{\star}$} & 43.29\% & 39.02\% & 41.40\% & 61.11\% & 53.90\% & 25.48\% & 44.03\% \\
 & & (6.50) & (6.50) & (9.25) & (9.43) & (5.07) & (7.92) & (7.45) \\
\midrule
\midrule
Qwen3 & 4B & 57.32\% & 50.61\% & 67.00\% & 80.69\% & 77.48\% & 47.10\% & 63.37\% \\
Qwen3 & 8B & 64.63\% & 56.71\% & 69.40\% & 83.07\% & 81.80\% & 52.94\% & 68.09\% \\
\midrule
LLaDA & 8B & 32.32\% & 27.44\% & 40.80\% & 51.85\% & 70.96\% & 27.30\% & 41.78\% \\
Dream & 7B & 54.88\% & 49.39\% & 56.80\% & 74.60\% & 77.18\% & 39.60\% & 58.74\% \\
Block Diff & 4B$^{\dagger}$ & 56.10\% & 51.22\% & 54.60\% & 69.84\% & 82.87\% & 47.02\% & 60.27\% \\
\multirow{2}{*}{\ours{} (Trust AR)} & \multirow{2}{*}{8B$^{\ddagger}$} & 55.49\% & 52.44\% & 65.40\% & 79.63\% & 79.83\% & 50.58\% & 63.90\% \\
 & & (7.46) & (7.44) & (9.96) & (10.13) & (6.90) & (7.48) & (8.23) \\
\multirow{2}{*}{\ours{} (Trust Diff)} & \multirow{2}{*}{8B$^{\ddagger}$} & 57.93\% & 55.49\% & 65.40\% & 80.95\% & 80.44\% & 51.64\% & 65.31\% \\
 & & (7.30) & (7.29) & (10.00) & (10.13) & (7.07) & (7.68) & (8.25) \\
\midrule
\end{tabular}
\caption{\textbf{Generative Evaluation Results:} We evaluate the generative sample quality of \ours{} over several coding and math tasks. For all diffusion models we decode one token per forward for the best quality and report the average number of tokens (T/NFE) generated by \ours{} in parenthesis. Models initialized from \textsc{Qwen2.5 1.5B}, \textsc{Qwen3 4B}, and \textsc{Qwen3 8B} are super-scripted by $\star$, $\dagger$, and $\ddagger$ respectively.}
\label{tab:generative}
\end{table}

\subsubsection{Likelihood Task Evaluation}
\label{sec:likelihood}
Evaluating model performance based on likelihood for traditional diffusion LLMs (e.g. LLaDA, Dream, MDLM) has been very challenging due to various ways of computing the likelihood. For example, response tokens will be corrupted in the same way as training, and the likelihood is calculated only on the masked positions~\cite{nie2025large}, which are averaged over Monte Carlo sample budgets. Although agreement has been achieved on the use of likelihood for multiple-choice questions despite being less efficient, directly comparing models using this metric still remains a subject of debate. \ours{}, on the other hand, alleviates this problem due to the fact that its AR mode naturally supports computing the likelihood in the same way as autoregressive models. We show in Table~\ref{tab:likelihood} that 1) this approach is highly aligned with other AR LLMs, making it highly comparable, 2) its performance is competitive and faithful to generative quality (due to autoregressive sampling), and 3) the evaluation is extremely efficient with a single NFE. 

\begin{table}[h]
\fontsize{8}{9}\selectfont
\centering
\begin{tabular}{llc|ccccc|c}
\toprule
 & & \multicolumn{1}{c}{\textbf{Knowledge}} & \multicolumn{5}{c}{\textbf{Commonsense Reasoning}} \\
\midrule
\textbf{Model Arch} & \textbf{Size} & \multicolumn{1}{c}{\textbf{MMLU}} & \multicolumn{1}{c}{\textbf{ARC-e}} & \multicolumn{1}{c}{\textbf{ARC-c}} & \multicolumn{1}{c}{\textbf{Hellaswag}} & \multicolumn{1}{c}{\textbf{PIQA}} & \multicolumn{1}{c}{\textbf{Winogrande}} & \multicolumn{1}{c}{\textbf{Avg}} \\
\midrule
Llama3.2 & 1B & 30.98\% & 65.28\% & 36.35\% & 63.76\% & 74.43\% & 63.30\% & 55.68\% \\
SmolLM2 & 1.7B & 49.99\% & 77.82\% & 47.44\% & 71.44\% & 77.64\% & 67.88\% & 65.37\% \\
Qwen2.5 & 0.5B & 47.65\% & 64.77\% & 31.83\% & 52.25\% & 70.02\% & 57.70\% & 54.04\% \\
Qwen2.5 & 1.5B & 60.96\% & 75.17\% & 45.05\% & 67.90\% & 76.12\% & 65.75\% & 65.16\% \\
Qwen3 & 1.7B & 62.53\% & 73.32\% & 44.62\% & 66.43\% & 75.63\% & 64.96\% & 64.58\% \\
\midrule
\textit{Block Diff} & 1.5B$^{\star}$ & 57.94\% & 74.41\% & 45.73\% & 56.26\% & 70.13\% & 61.80\% & 61.05\% \\
\ours{} & 1.5B$^{\star}$ & 58.99\% & 77.78\% & 45.39\% & 65.26\% & 75.52\% & 63.61\% & 64.43\% \\
\midrule
\midrule
Qwen3 & 4B & 73.00\% & 79.00\% & 52.00\% & 74.00\% & 78.00\% & 72.00\% & 71.33\% \\
Qwen3 & 8B & 76.93\% & 81.90\% & 53.16\% & 78.59\% & 79.22\% & 75.69\% & 74.25\% \\
\midrule
\textit{Block Diff} & 4B$^{\dagger}$ & 71.53\% & 81.48\% & 55.63\% & 65.48\% & 74.92\% & 70.96\% & 70.00\% \\
\textit{Dream} & 7B & 67.00\% & 82.20\% & 59.13\% & 73.73\% & 75.52\% & 73.56\% & 71.86\% \\
\textit{LLaDA} & 8B & 65.86\% & 73.78\% & 49.15\% & 71.05\% & 73.88\% & 74.66\% & 68.06\% \\
\ours{} & 8B$^{\ddagger}$ & 76.57\% & 84.18\% & 58.53\% & 76.36\% & 80.25\% & 76.48\% & 75.40\% \\
\bottomrule
\end{tabular}
\caption{\textbf{Likelihood Evaluation Results:} We compare our method on factual knowledge and common sense reasoning tasks using likelihood estimation. For Llada, Dream, and Block Diffusion, we follow the standard diffusion likelihood evaluation using Monte Carlo (MC) sampling. For our model, we evaluate the likelihood using pure causal mask like AR models, as natively supported by our architecture. Models initialized from \textsc{Qwen2.5 1.5B}, \textsc{Qwen3 4B}, and \textsc{Qwen3 8B} are super-scripted by $\star$, $\dagger$, and $\ddagger$ respectively. Models with \textit{italicized} names are using MC with 128 steps for likelihood evaluation. }
\label{tab:likelihood}
\end{table}

\subsection{Efficiency Benchmarking}
Finally, we conclude the main results with a focus on benchmarking the wall-clock time of \ours{} against AR, AR with EAGLE-3\footnote{We use the EAGLE-3 with Qwen3-8B instruct model due to lack of corresponding EAGLE-3 weights for the base model.} for speculative decoding, and Block Diffusion. We choose these since they are among the most competitive choices across standard autoregressive decoding, speculative decoding, and diffusion generations. All of the models have the exact cache supported and are benchmarked on a single H100 GPU with prompts from downstream generative tasks and a batch size equal to 1. 

In Figure~\ref{fig:throughput}, we show that \ours{} 1.5B can achieve an average of 4.71x relative speedup to Qwen2.5 1.5B in terms of decoding throughput and \ours{} 8B with an average of 5.91x over Qwen3 8B while maintaining comparable performances. Comparing with Block Diffusion with two settings using different thresholds, we can also achieve better efficiency-quality trade-offs in all tasks we tested. In terms of the SOTA speculative decoding method, EAGLE-3~\cite{li2025eagle3}, we test \ours{} against public weights from AngelSlim\footnote{\url{https://huggingface.co/AngelSlim/Qwen3-8B_eagle3}} and Tengyunw\footnote{\url{https://huggingface.co/Tengyunw/qwen3_8b_eagle3}}, in which we show for the first time that diffusion models can surpass the efficiency gains over speculative decoding. However, we want to emphasize that speculative decoding and our methods should be utilized with different purposes since the former is able to guarantee exactly the same output as the base model. In addition, we show that our raw acceptance rate (T/NFE) is higher than those of EAGLE-3 open weights, and more importantly, our conversion rate (from T/NFE to T/s) is higher, thanks to the parallel drafting and sampling with a single model forward. 

It is worth noting that all of these methods can significantly benefit from further system optimizations such as custom kernels, more efficient KV cache management, and request scheduling. Hence, the goal here is to provide a first-hand idea of their performance in native PyTorch. 

\subsection{Ablation Studies}

The objective of this section is to understand more deeply where the performance gains come from and justify the design choices of our architecture, training, and inference. 

\subsubsection{Pareto Frontier under the Same Training Recipe}

\begin{figure}[h]
    \centering
    \includegraphics[width=0.8\columnwidth]{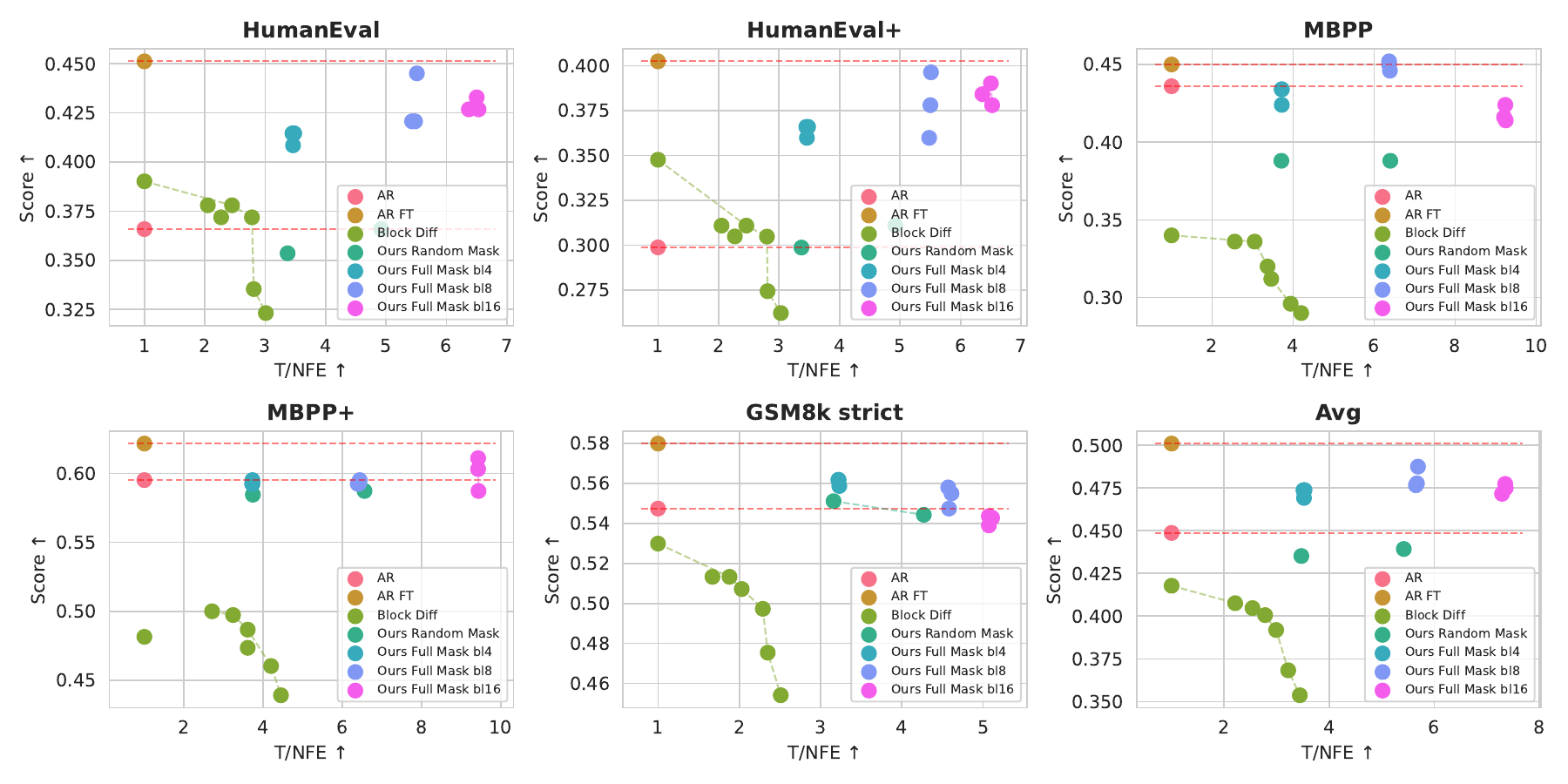}
    \caption{\textbf{Pareto Frontier of Different Architectures with the Same Recipe:} We report the performance-efficiency trade-offs on 1.5B scale among AR model, fine-tuned AR model, Block Diffusion under different decoding thresholds, and \ours{} using different drafting lengths. Our model achieves the best Pareto Frontier compared to Block Diffusion and AR and is approaching the quality of fine-tuned AR with 7x more tokens per NFE. }
    \label{fig:pareto_1p5b}
\end{figure}

The quality and training recipe can drastically impact the performance of LLMs. Therefore, to fairly compare different model architectures, we train all models using the same setting starting from Qwen2.5 1.5B models. In Figure~\ref{fig:pareto_1p5b}, the Pareto Frontier of these trained models provides insight into how \ours{} performs against the base AR, Block Diffusion, and fine-tuned AR models. We can see that with only 50B training tokens, \ours{} can achieve impressive quality while maintaining a high T/NFE compared with Block Diffusion with different thresholds used for parallel decoding. We also notice the remaining quality gap from the fine-tuned AR models, which we believe could be potentially closed with more data due to the fact that \ours{} might require a bit more knowledge due to the initial adaptation phase. 

\subsubsection{Comparing \ours{} with Other Decoding Strategy}
\label{sec:compare_decoding}

\begin{table}[h]
\fontsize{8}{9}\selectfont
\centering
\begin{tabular}{lcccc}
\toprule
\textbf{Strategy} & \multicolumn{1}{c}{\textbf{Avg T/NFE}} & \multicolumn{1}{c}{\textbf{HumanEval Avg}} & \multicolumn{1}{c}{\textbf{MBPP Avg}} & \multicolumn{1}{c}{\textbf{GSM8k}} \\
\midrule
\multirow{2}{*}{Confidence Max} & 1.00 & 34.45\% & 43.92\% & 53.07\% \\
 & 2.00 & 23.78\% & 22.19\% & 38.06\% \\
\midrule
\multirow{2}{*}{Left to right (AR)} & 1.00 & 36.28\% & 46.51\% & 53.37\% \\
 & 2.00 & 21.95\% & 18.61\% & 41.32\% \\
\midrule
Confidence \textgreater{} 0.9 & 2.63 & 32.01\% & 42.50\% & 51.40\% \\
Confidence \textgreater{} 0.8 & 3.06 & 28.96\% & 39.28\% & 47.54\% \\
Confidence \textgreater{} 0.7 & 3.42 & 27.74\% & 33.90\% & 43.44\% \\
Confidence \textgreater{} 0.6 & 3.81 & 22.56\% & 26.47\% & 37.60\% \\
\midrule
\ours{} (4 drafts) & 3.47 & 38.42\% & 50.96\% & 55.87\% \\
\ours{} (8 drafts) & 5.49 & 39.94\% & 52.13\% & 54.74\% \\
\ours{} (16 drafts) & 6.97 & 41.16\% & 51.26\% & 53.90\% \\
\bottomrule
\end{tabular}
\caption{\textbf{Comparing Different Decoding Strategies:} We showcase the superiority of our proposed parallel draft and sampling process (on full mask models) compared to using standard confidence/negative entropy based decoding, as well as the autoregressive decoding schemes. }
\label{tab:decoding}
\end{table}

Common strategies used in dLMs include generating a fixed number of tokens per NFE~\cite{nie2025large}, dynamic number of NFEs based on entropy~\cite{ye2025dream} or confidence-based decoding~\cite{wu2025fastdllmtrainingfreeaccelerationdiffusion}, and also within-block left-to-right decoding. Many prior works have studied the benefits of these and, here, we provide a comprehensive comparison in Table~\ref{tab:decoding} against our method. We demonstrate that our method leverages both the efficient parallelism from diffusion but also high quality from autoregression, with an added benefit of not requiring to tune any hyperparameters during decoding. 

\subsubsection{Sampling with AR v.s. Diffusion Prediction}
\label{sec:sampling_ar_vs_diff}

\begin{figure}[h]
    \centering
    \includegraphics[width=0.5\columnwidth]{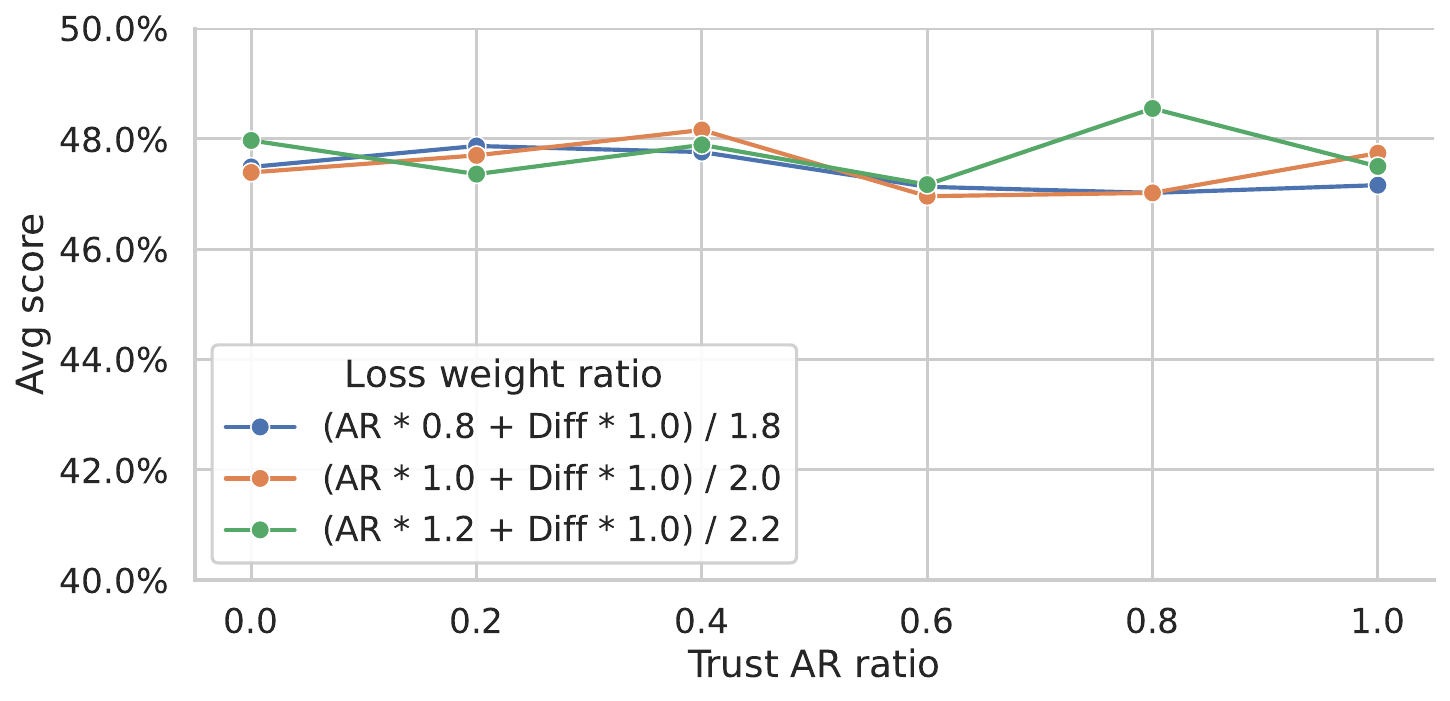}
    \caption{\textbf{Trusting AR v.s. Diffusion Outputs for Sampling:} \ours{} is trained to be highly balanced in the sense that no matter whether we trust the prediction from the diffusion or AR parts, the autoregressive sampling, along with the high drafting model capacity, can guarantee quality under almost the same speedup. }
    \label{fig:trust_ar_vs_diff}
\end{figure}

As shown in Figure~\ref{fig:main}, using no label shifts for the diffusion part makes it possible to pre-draft without waiting for AR's results at the same step while keeping the sequence consistent (i.e. the first mask would predict $F''$ instead without knowing the actual value of $E$ when it is label shifted, which will degrade the quality). However, this comes at the cost that the AR outputs and every first position in the block will be predicting the same position (e.g. $E$ to $E''$, $F'$ to $F''$, $G'$ to $G''$), leading to a potential decision of choosing what to use for verification. 

Therefore, we propose to apply an aggregate function to mix their logits before sampling the token as $\arg\max_i \{\beta * logits^{ar}_i + (1 - \beta) * logits^{diff}_i, i \in |V|\}$, which then gets compared against the proposed draft token. This can also be understood as whether we ``trust'' the AR or the diffusion outputs more when it comes to sampling. 

In Figure~\ref{fig:trust_ar_vs_diff}, we vary the value of $\beta$ (over different loss balancing factor $\alpha$) and see consistent performance. This shows that our model is well trained so that no matter what logits we choose to use for verification, the quality is preserved because in the ideal case (i.e. well-trained), these two outputs will be strictly the same. This also indicates that it is the autoregressive rejection sampling that guarantees the quality-speedup trade-offs rather than the AR knowledge.

\subsubsection{How Useful is the Full Mask Strategy?}
\label{sec:full_mask}
To take advantage of the fact that the model does not see partial clean inputs during inference due to one-step diffusion drafting, we corrupt the inputs to the diffusion part to be full masks during training. This not only promotes better train-test behavior consistency but also provides richer diffusion loss signals and easier loss balancing after normalization. In Table~\ref{tab:full_mask}, we demonstrate the effectiveness of this masking strategy, which shows a consistent quality improvement especially in coding tasks. We also want to note that this enables the flexibility of deciding which module's predictions to trust for autoregressive rejection sampling as discussed in Section~\ref{sec:sampling_ar_vs_diff}. 

\begin{table}[h]
\fontsize{8}{9}\selectfont
\centering
\begin{tabular}{lcccccc}
\toprule
\textbf{Masking} & \textbf{Draft} & \multicolumn{1}{c}{\textbf{HumanEval Avg}} & \multicolumn{1}{c}{\textbf{MBPP Avg}} & \multicolumn{1}{c}{\textbf{GSM8k}} & \multicolumn{1}{c}{\textbf{Avg}} \\
\midrule
\multirow{4}{*}{Random} & \multirow{2}{*}{4} & \multicolumn{1}{c}{32.62\%} & \multicolumn{1}{c}{48.63\%} & \multicolumn{1}{c}{55.11\%} & \multicolumn{1}{c}{45.45\%} \\
 &  & (3.37) & (3.72) & (3.16) & (3.42) \\
 & \multirow{2}{*}{8} & \multicolumn{1}{c}{33.85\%} & \multicolumn{1}{c}{48.77\%} & \multicolumn{1}{c}{54.43\%} & \multicolumn{1}{c}{45.68\%} \\
 &  & (4.92) & (6.48) & (4.27) & (5.22) \\
\midrule
\multirow{4}{*}{Full} & \multirow{2}{*}{4} & \multicolumn{1}{c}{38.42\%} & \multicolumn{1}{c}{50.96\%} & \multicolumn{1}{c}{55.87\%} & \multicolumn{1}{c}{48.42\%} \\
 &  & (3.46) & (3.72) & (3.23) & (3.47) \\
 & \multirow{2}{*}{8} & \multicolumn{1}{c}{39.94\%} & \multicolumn{1}{c}{52.13\%} & \multicolumn{1}{c}{54.74\%} & \multicolumn{1}{c}{48.94\%} \\
 &  & (5.49) & (6.40) & (4.58) & (5.49) \\
\bottomrule
\end{tabular}
\caption{\textbf{Quality-efficiency Improvement from Full Masking:} Turning random corruption strategy to full masking on the model, we substantially improve both efficiency and quality due to less train-test discrepancy and richer diffusion loss signals. Average T/NFE is shown in the parenthesis. }
\label{tab:full_mask}
\end{table}

\section{Limitations}
\label{limitations}

\paragraph{Batch size} Although we focus on batch size = 1 efficiency benchmarking, it does not mean that \ours{} cannot handle large batch size. Not only can we adjust the block (draft) length during decoding in a zero-shot manner to accommodate different compute profile, but also can achieve competitive performance in terms of FLOPs / token. 

\paragraph{Long context extension} We believe that our model is not intrinsically limited in long context capability compared to standard AR models. Since our current implementation requires doubling the sequence length with appended mask tokens during training, we defer the exploration of efficient long context extension methods (e.g. context parallelism specifically designed for \ours{}) for future work. 

\paragraph{System optimization} We show that even writing in native PyTorch with Flex Attention, we have already achieved substantial throughput improvement. We believe that writing custom attention kernels and scheduling algorithms can maximize the use of the ``free token slots'' specific to the serving hardware in use as shown in Figure~\ref{fig:latency_scaling}. 
\section{Conclusions}
\label{conclusions}

We introduce \ours{}, a sequence-level hybrid architecture that drafts (thinks) in diffusion and samples (talks) in autoregression in a single model forward with a specially designed attention mask. Taking advantage of the efficiency gains from parallel one-step diffusion and the quality guaranteed by autoregressive sampling, \ours{} achieves impressive efficiency and quality trade-offs. We show in a diverse set of downstream tasks that \ours{} 1.5B and 8B models can output with an average of 7.45 and 8.25 tokens per NFE, which translates to $4.71\times$ and $5.91\times$ more tokens per second compared to AR while maintaining competitive quality. We push the limit of ``free token slots'' on modern GPUs and, for the first time, show it is possible to beat speculative decoding for latency-critical applications. We believe that this will significantly motivate future research on hybrid LLM architecture and inference.

\newpage
{
  \small
  \bibliographystyle{unsrt}
  \bibliography{main_tech_report}

\begin{thebibliography}{10}

\bibitem{bubeck2023sparks}
S{\'e}bastien Bubeck, Varun Chandrasekaran, Ronen Eldan, Johannes Gehrke, Eric Horvitz, Ece Kamar, Peter Lee, Yin~Tat Lee, Yuanzhi Li, Scott Lundberg, et~al.
\newblock Sparks of artificial general intelligence: Early experiments with gpt-4.
\newblock {\em arXiv preprint arXiv:2303.12712}, 2023.

\bibitem{nvidia_gpu}
NVIDIA.
\newblock Nvidia gpus, 2025.
\newblock General-purpose GPU.

\bibitem{vaswani2017attention}
Ashish Vaswani, Noam Shazeer, Niki Parmar, Jakob Uszkoreit, Llion Jones, Aidan~N Gomez, {\L}ukasz Kaiser, and Illia Polosukhin.
\newblock Attention is all you need.
\newblock {\em Advances in neural information processing systems}, 30, 2017.

\bibitem{radford2019language}
Alec Radford, Jeff Wu, Rewon Child, David Luan, Dario Amodei, and Ilya Sutskever.
\newblock Language models are unsupervised multitask learners.
\newblock 2019.

\bibitem{dao2022flashattention}
Tri Dao, Dan Fu, Stefano Ermon, Atri Rudra, and Christopher R{\'e}.
\newblock Flashattention: Fast and memory-efficient exact attention with io-awareness.
\newblock {\em Advances in neural information processing systems}, 35:16344--16359, 2022.

\bibitem{yuan2024llm}
Zhihang Yuan, Yuzhang Shang, Yang Zhou, Zhen Dong, Zhe Zhou, Chenhao Xue, Bingzhe Wu, Zhikai Li, Qingyi Gu, Yong~Jae Lee, et~al.
\newblock Llm inference unveiled: Survey and roofline model insights.
\newblock {\em arXiv preprint arXiv:2402.16363}, 2024.

\bibitem{sahoo2024simpleeffectivemaskeddiffusion}
Subham~Sekhar Sahoo, Marianne Arriola, Yair Schiff, Aaron Gokaslan, Edgar Marroquin, Justin~T Chiu, Alexander Rush, and Volodymyr Kuleshov.
\newblock Simple and effective masked diffusion language models, 2024.

\bibitem{nie2025large}
Shen Nie, Fengqi Zhu, Zebin You, Xiaolu Zhang, Jingyang Ou, Jun Hu, Jun Zhou, Yankai Lin, Ji-Rong Wen, and Chongxuan Li.
\newblock Large language diffusion models.
\newblock {\em arXiv preprint arXiv:2502.09992}, 2025.

\bibitem{ye2025dream}
Jiacheng Ye, Zhihui Xie, Lin Zheng, Jiahui Gao, Zirui Wu, Xin Jiang, Zhenguo Li, and Lingpeng Kong.
\newblock Dream 7b: Diffusion large language models.
\newblock {\em arXiv preprint arXiv:2508.15487}, 2025.

\bibitem{kwon2023efficient}
Woosuk Kwon, Zhuohan Li, Siyuan Zhuang, Ying Sheng, Lianmin Zheng, Cody~Hao Yu, Joseph Gonzalez, Hao Zhang, and Ion Stoica.
\newblock Efficient memory management for large language model serving with pagedattention.
\newblock In {\em Proceedings of the 29th symposium on operating systems principles}, pages 611--626, 2023.

\bibitem{leviathan2023fast}
Yaniv Leviathan, Matan Kalman, and Yossi Matias.
\newblock Fast inference from transformers via speculative decoding.
\newblock In {\em International Conference on Machine Learning}, pages 19274--19286. PMLR, 2023.

\bibitem{arriola2025blockdiffusion}
Marianne Arriola, Aaron Gokaslan, Justin~T Chiu, Zhihan Yang, Zhixuan Qi, Jiaqi Han, Subham~Sekhar Sahoo, and Volodymyr Kuleshov.
\newblock Block diffusion: Interpolating between autoregressive and diffusion language models.
\newblock {\em arXiv preprint arXiv:2503.09573}, 2025.

\bibitem{wu2025fastdllmtrainingfreeaccelerationdiffusion}
Chengyue Wu, Hao Zhang, Shuchen Xue, Zhijian Liu, Shizhe Diao, Ligeng Zhu, Ping Luo, Song Han, and Enze Xie.
\newblock Fast-dllm: Training-free acceleration of diffusion llm by enabling kv cache and parallel decoding, 2025.

\bibitem{feng2025theoretical}
Guhao Feng, Yihan Geng, Jian Guan, Wei Wu, Liwei Wang, and Di~He.
\newblock Theoretical benefit and limitation of diffusion language model.
\newblock {\em arXiv preprint arXiv:2502.09622}, 2025.

\bibitem{dao2023flashattention2fasterattentionbetter}
Tri Dao.
\newblock Flashattention-2: Faster attention with better parallelism and work partitioning, 2023.

\bibitem{chen2023accelerating}
Charlie Chen, Sebastian Borgeaud, Geoffrey Irving, Jean-Baptiste Lespiau, Laurent Sifre, and John Jumper.
\newblock Accelerating large language model decoding with speculative sampling.
\newblock {\em arXiv preprint arXiv:2302.01318}, 2023.

\bibitem{israel2025apd}
Daniel Israel, Guy Van~den Broeck, and Aditya Grover.
\newblock Accelerating diffusion llms via adaptive parallel decoding.
\newblock {\em arXiv preprint arXiv:2506.00413}, 2025.

\bibitem{li2025eagle3}
Yuhui Li, Fangyun Wei, Chao Zhang, and Hongyang Zhang.
\newblock Eagle-3: Scaling up inference acceleration of large language models via training-time test.
\newblock {\em arXiv preprint arXiv:2503.01840}, 2025.

\bibitem{liu2024deepseekv3}
Aixin Liu, Bei Feng, Bing Xue, Bingxuan Wang, Bochao Wu, Chengda Lu, Chenggang Zhao, Chengqi Deng, Chenyu Zhang, Chong Ruan, et~al.
\newblock Deepseek-v3 technical report.
\newblock {\em arXiv preprint arXiv:2412.19437}, 2024.

\bibitem{samragh2025llmknowsfutureuncovering}
Mohammad Samragh, Arnav Kundu, David Harrison, Kumari Nishu, Devang Naik, Minsik Cho, and Mehrdad Farajtabar.
\newblock Your llm knows the future: Uncovering its multi-token prediction potential, 2025.

\bibitem{austin2021structured}
Jacob Austin, Daniel~D Johnson, Jonathan Ho, Daniel Tarlow, and Rianne Van Den~Berg.
\newblock Structured denoising diffusion models in discrete state-spaces.
\newblock {\em Advances in neural information processing systems}, 34:17981--17993, 2021.

\bibitem{li2022diffusion}
Xiang Li, John Thickstun, Ishaan Gulrajani, Percy~S Liang, and Tatsunori~B Hashimoto.
\newblock Diffusion-lm improves controllable text generation.
\newblock {\em Advances in neural information processing systems}, 35:4328--4343, 2022.

\bibitem{cobbe2021training}
Karl Cobbe, Vineet Kosaraju, Mohammad Bavarian, Mark Chen, Heewoo Jun, Lukasz Kaiser, Matthias Plappert, Jerry Tworek, Jacob Hilton, Reiichiro Nakano, et~al.
\newblock Training verifiers to solve math word problems.
\newblock {\em arXiv preprint arXiv:2110.14168}, 2021.

\bibitem{wu2025fastdllmv2efficientblockdiffusion}
Chengyue Wu, Hao Zhang, Shuchen Xue, Shizhe Diao, Yonggan Fu, Zhijian Liu, Pavlo Molchanov, Ping Luo, Song Han, and Enze Xie.
\newblock Fast-dllm v2: Efficient block-diffusion llm, 2025.

\bibitem{arriola2025encoderdecoderdiffusionlanguagemodels}
Marianne Arriola, Yair Schiff, Hao Phung, Aaron Gokaslan, and Volodymyr Kuleshov.
\newblock Encoder-decoder diffusion language models for efficient training and inference, 2025.

\bibitem{xu2025energybaseddiffusionlanguagemodels}
Minkai Xu, Tomas Geffner, Karsten Kreis, Weili Nie, Yilun Xu, Jure Leskovec, Stefano Ermon, and Arash Vahdat.
\newblock Energy-based diffusion language models for text generation, 2025.

\bibitem{ma2025dkvcachecachediffusionlanguage}
Xinyin Ma, Runpeng Yu, Gongfan Fang, and Xinchao Wang.
\newblock dkv-cache: The cache for diffusion language models, 2025.

\bibitem{cai2024medusa}
Tianle Cai, Yuhong Li, Zhengyang Geng, Hongwu Peng, Jason~D Lee, Deming Chen, and Tri Dao.
\newblock Medusa: Simple llm inference acceleration framework with multiple decoding heads.
\newblock {\em arXiv preprint arXiv:2401.10774}, 2024.

\bibitem{li2024eagle1}
Yuhui Li, Fangyun Wei, Chao Zhang, and Hongyang Zhang.
\newblock Eagle: Speculative sampling requires rethinking feature uncertainty.
\newblock {\em arXiv preprint arXiv:2401.15077}, 2024.

\bibitem{li2024eagle2}
Yuhui Li, Fangyun Wei, Chao Zhang, and Hongyang Zhang.
\newblock Eagle-2: Faster inference of language models with dynamic draft trees.
\newblock {\em arXiv preprint arXiv:2406.16858}, 2024.

\bibitem{gat2025set}
Itai Gat, Heli Ben-Hamu, Marton Havasi, Daniel Haziza, Jeremy Reizenstein, Gabriel Synnaeve, David Lopez-Paz, Brian Karrer, and Yaron Lipman.
\newblock Set block decoding is a language model inference accelerator.
\newblock {\em arXiv preprint arXiv:2509.04185}, 2025.

\bibitem{sahoo2024simple}
Subham Sahoo, Marianne Arriola, Yair Schiff, Aaron Gokaslan, Edgar Marroquin, Justin Chiu, Alexander Rush, and Volodymyr Kuleshov.
\newblock Simple and effective masked diffusion language models.
\newblock {\em Advances in Neural Information Processing Systems}, 37:130136--130184, 2024.

\bibitem{gloeckle2024better}
Fabian Gloeckle, Badr~Youbi Idrissi, Baptiste Rozi{\`e}re, David Lopez-Paz, and Gabriel Synnaeve.
\newblock Better \& faster large language models via multi-token prediction.
\newblock {\em arXiv preprint arXiv:2404.19737}, 2024.

\bibitem{liu2025sequential}
Yangzhou Liu, Yue Cao, Hao Li, Gen Luo, Zhe Chen, Weiyun Wang, Xiaobo Liang, Biqing Qi, Lijun Wu, Changyao Tian, et~al.
\newblock Sequential diffusion language models.
\newblock {\em arXiv preprint arXiv:2509.24007}, 2025.

\bibitem{dong2024flex}
Juechu Dong, Boyuan Feng, Driss Guessous, Yanbo Liang, and Horace He.
\newblock Flex attention: A programming model for generating optimized attention kernels.
\newblock {\em arXiv preprint arXiv:2412.05496}, 2024.

\bibitem{qwen2025qwen25technicalreport}
An~Yang, Baosong Yang, Beichen Zhang, Binyuan Hui, Bo~Zheng, Bowen Yu, Chengyuan Li, Dayiheng Liu, Fei Huang, Haoran Wei, Huan Lin, Jian Yang, Jianhong Tu, Jianwei Zhang, Jianxin Yang, Jiaxi Yang, Jingren Zhou, Junyang Lin, Kai Dang, Keming Lu, Keqin Bao, Kexin Yang, Le~Yu, Mei Li, Mingfeng Xue, Pei Zhang, Qin Zhu, Rui Men, Runji Lin, Tianhao Li, Tianyi Tang, Tingyu Xia, Xingzhang Ren, Xuancheng Ren, Yang Fan, Yang Su, Yichang Zhang, Yu~Wan, Yuqiong Liu, Zeyu Cui, Zhenru Zhang, and Zihan Qiu.
\newblock Qwen2.5 technical report, 2025.

\bibitem{yang2025qwen3technicalreport}
An~Yang, Anfeng Li, Baosong Yang, Beichen Zhang, Binyuan Hui, Bo~Zheng, Bowen Yu, Chang Gao, Chengen Huang, Chenxu Lv, Chujie Zheng, Dayiheng Liu, Fan Zhou, Fei Huang, Feng Hu, Hao Ge, Haoran Wei, Huan Lin, Jialong Tang, Jian Yang, Jianhong Tu, Jianwei Zhang, Jianxin Yang, Jiaxi Yang, Jing Zhou, Jingren Zhou, Junyang Lin, Kai Dang, Keqin Bao, Kexin Yang, Le~Yu, Lianghao Deng, Mei Li, Mingfeng Xue, Mingze Li, Pei Zhang, Peng Wang, Qin Zhu, Rui Men, Ruize Gao, Shixuan Liu, Shuang Luo, Tianhao Li, Tianyi Tang, Wenbiao Yin, Xingzhang Ren, Xinyu Wang, Xinyu Zhang, Xuancheng Ren, Yang Fan, Yang Su, Yichang Zhang, Yinger Zhang, Yu~Wan, Yuqiong Liu, Zekun Wang, Zeyu Cui, Zhenru Zhang, Zhipeng Zhou, and Zihan Qiu.
\newblock Qwen3 technical report, 2025.

\bibitem{eval-harness}
Leo Gao, Jonathan Tow, Baber Abbasi, Stella Biderman, Sid Black, Anthony DiPofi, Charles Foster, Laurence Golding, Jeffrey Hsu, Alain Le~Noac'h, Haonan Li, Kyle McDonell, Niklas Muennighoff, Chris Ociepa, Jason Phang, Laria Reynolds, Hailey Schoelkopf, Aviya Skowron, Lintang Sutawika, Eric Tang, Anish Thite, Ben Wang, Kevin Wang, and Andy Zou.
\newblock The language model evaluation harness, 07 2024.

\bibitem{meta2024llama}
AI~Meta.
\newblock Llama 3.2: Revolutionizing edge ai and vision with open, customizable models.
\newblock {\em Meta AI Blog. Retrieved December}, 20:2024, 2024.

\bibitem{allal2025smollm2}
Loubna~Ben Allal, Anton Lozhkov, Elie Bakouch, Gabriel~Mart{\'\i}n Bl{\'a}zquez, Guilherme Penedo, Lewis Tunstall, Andr{\'e}s Marafioti, Hynek Kydl{\'\i}{\v{c}}ek, Agust{\'\i}n~Piqueres Lajar{\'\i}n, Vaibhav Srivastav, et~al.
\newblock Smollm2: When smol goes big--data-centric training of a small language model.
\newblock {\em arXiv preprint arXiv:2502.02737}, 2025.

\bibitem{kingma2017adammethodstochasticoptimization}
Diederik~P. Kingma and Jimmy Ba.
\newblock Adam: A method for stochastic optimization, 2017.

\bibitem{shoeybi2020megatronlmtrainingmultibillionparameter}
Mohammad Shoeybi, Mostofa Patwary, Raul Puri, Patrick LeGresley, Jared Casper, and Bryan Catanzaro.
\newblock Megatron-lm: Training multi-billion parameter language models using model parallelism, 2020.

\bibitem{liang2025torchtitanonestoppytorchnative}
Wanchao Liang, Tianyu Liu, Less Wright, Will Constable, Andrew Gu, Chien-Chin Huang, Iris Zhang, Wei Feng, Howard Huang, Junjie Wang, Sanket Purandare, Gokul Nadathur, and Stratos Idreos.
\newblock Torchtitan: One-stop pytorch native solution for production ready llm pre-training, 2025.

\bibitem{chen2021evaluating}
Mark Chen, Jerry Tworek, Heewoo Jun, Qiming Yuan, Henrique Ponde De~Oliveira Pinto, Jared Kaplan, Harri Edwards, Yuri Burda, Nicholas Joseph, Greg Brockman, et~al.
\newblock Evaluating large language models trained on code.
\newblock {\em arXiv preprint arXiv:2107.03374}, 2021.

\bibitem{liu2023your}
Jiawei Liu, Chunqiu~Steven Xia, Yuyao Wang, and Lingming Zhang.
\newblock Is your code generated by chatgpt really correct? rigorous evaluation of large language models for code generation.
\newblock {\em Advances in Neural Information Processing Systems}, 36:21558--21572, 2023.

\bibitem{austin2021program}
Jacob Austin, Augustus Odena, Maxwell Nye, Maarten Bosma, Henryk Michalewski, David Dohan, Ellen Jiang, Carrie Cai, Michael Terry, Quoc Le, et~al.
\newblock Program synthesis with large language models.
\newblock {\em arXiv preprint arXiv:2108.07732}, 2021.

\bibitem{lewkowycz2022solving}
Aitor Lewkowycz, Anders Andreassen, David Dohan, Ethan Dyer, Henryk Michalewski, Vinay Ramasesh, Ambrose Slone, Cem Anil, Imanol Schlag, Theo Gutman-Solo, et~al.
\newblock Solving quantitative reasoning problems with language models.
\newblock {\em Advances in neural information processing systems}, 35:3843--3857, 2022.

\bibitem{hendrycks2020measuring}
Dan Hendrycks, Collin Burns, Steven Basart, Andy Zou, Mantas Mazeika, Dawn Song, and Jacob Steinhardt.
\newblock Measuring massive multitask language understanding.
\newblock {\em arXiv preprint arXiv:2009.03300}, 2020.

\bibitem{clark2018think}
Peter Clark, Isaac Cowhey, Oren Etzioni, Tushar Khot, Ashish Sabharwal, Carissa Schoenick, and Oyvind Tafjord.
\newblock Think you have solved question answering? try arc, the ai2 reasoning challenge.
\newblock {\em arXiv preprint arXiv:1803.05457}, 2018.

\bibitem{zellers2019hellaswag}
Rowan Zellers, Ari Holtzman, Yonatan Bisk, Ali Farhadi, and Yejin Choi.
\newblock Hellaswag: Can a machine really finish your sentence?
\newblock {\em arXiv preprint arXiv:1905.07830}, 2019.

\bibitem{bisk2020piqa}
Yonatan Bisk, Rowan Zellers, Jianfeng Gao, Yejin Choi, et~al.
\newblock Piqa: Reasoning about physical commonsense in natural language.
\newblock In {\em Proceedings of the AAAI conference on artificial intelligence}, volume~34, pages 7432--7439, 2020.

\bibitem{sakaguchi2021winogrande}
Keisuke Sakaguchi, Ronan~Le Bras, Chandra Bhagavatula, and Yejin Choi.
\newblock Winogrande: An adversarial winograd schema challenge at scale.
\newblock {\em Communications of the ACM}, 64(9):99--106, 2021.

\end{thebibliography}
}

\newpage
\appendix
\section{Evaluation Task Configuration}
\label{sec:app_eval_task_config}

\begin{table}[h]
\centering
\fontsize{8}{9}\selectfont
\begin{tabular}{llccc}
\toprule
\multirow{2}{*}{\textbf{Task Category}} & \multirow{2}{*}{\textbf{Task Name}} & \textbf{Number of} & \textbf{Gen} & \textbf{Score} \\
&  & \textbf{Few-Shots} & \textbf{Length} & \textbf{Metric} \\
\midrule
\multirow{4}{*}{Coding} 
& HumanEval~\cite{chen2021evaluating} & 0 & 512 & Pass@1 \\
& HumanEval+~\cite{liu2023your} & 0 & 512 & Pass@1 \\
& MBPP~\cite{austin2021program} & 3 & 512 & Pass@1 \\
& MBPP+~\cite{liu2023your} & 3 & 512 & Pass@1 \\
\midrule
\multirow{2}{*}{Math} 
& GSM8K-CoT~\cite{cobbe2021training} & 8 & 256 & strict-match \\
& Minerva Math~\cite{lewkowycz2022solving} & 4 & 512 & exact-match \\
\midrule
Factual 
& MMLU~\cite{hendrycks2020measuring} & 5 & / & Acc \\
\midrule
& ARC-Easy~\cite{clark2018think} & 0 & / & Acc \\
& ARC-Challenge~\cite{clark2018think} & 0 & / & Acc Norm \\
Reasoning& Hellaswag~\cite{zellers2019hellaswag} & 0 & / & Acc Norm \\
& PIQA~\cite{bisk2020piqa} & 0 & / & Acc Norm \\
& Winogrande~\cite{sakaguchi2021winogrande} & 5 & / & Acc \\
\bottomrule
\end{tabular}
\caption{\textbf{Benchmark Configuration for All Tasks}}
\label{tab:detailed_task_setting}
\end{table}

In Tab.~\ref{tab:detailed_task_setting}, we detail the benchmarking configuration, including the number of few-shot prompts, generation length and performance metric, for each of tasks we adopted in the work. For HumanEval and HumanEval Plus, we apply standard post-processing for the all models. The \textsc{lm\_eval\_harness} version we use is 0.4.8. 

\section{Inference Prefill Mask}
\label{sec:app_prefill_mask}

\begin{figure}[h]
    \centering
    \includegraphics[width=0.6\columnwidth]{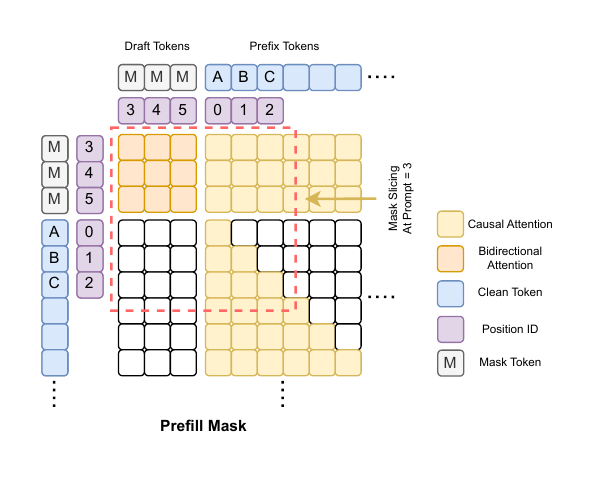}
    \vspace{-0.5cm}
    \caption{\textbf{Prefill Attention Mask:} We initialize the mask at the model initialization with (max\_seq\_len + block\_size, max\_seq\_len + block\_size) and slice it based on the current sample length. }
    \label{fig:prefill}
\end{figure}

We apply the same technique of reusing the initialized attention mask for different prompt lengths across samples. This is achieved by reordering different parts of the input and slicing the big mask. The mask is illustrated in Figure~\ref{fig:prefill}. 

\end{document}